\documentclass[]{Liebert_Author}

\usepackage{siunitx}
\DeclareSIUnit{\fahrenheit}{F}

\usepackage[normalem]{ulem}

\title{Design of a bioinspired robophysical antenna for insect-scale tactile perception and navigation} 

\author{$^{1}$Parker McDonnell, $^{2}$Lingsheng Meng, $^{1,2}$Hari Krishna Hariprasad, \\
$^{1}$Alexander Hedrick, $^{1}$Eduardo Miscles, $^{1}$ Samuel Gilinsky,  
\\
$^{2}$Jean-Michel Mongeau, $^{1}$Kaushik Jayaram$^{\ast}$\\
{$^{1}$Department of Mechanical Engineering, University of Colorado Boulder,}\\
{1111 Engineering Drive, Boulder, CO 80301, USA}\\
{$^{2}$Department of Mechanical Engineering, Penn State University}\\
{$^\ast$To whom correspondence should be addressed;}\\
{E-mail:  kaushik.jayaram@colorado.edu.}}

\begin{document} 

\maketitle
\keywords{tactile sensor, capacitive sensing and robophysical antenna}


\begin{abstract}

The American cockroach (\textit{Periplaneta americana}) uses its soft antennae to guide decision making by extracting rich tactile information from tens of thousands of distributed mechanosensors. Although tactile sensors enable robust, autonomous perception and navigation in natural systems, replicating these capabilities in insect-scale robots remains challenging due to stringent size, weight, and power constraints that limit existing sensor technologies. To overcome these limitations, we introduce CITRAS (Cockroach Inspired Tactile Robotic Antenna Sensor), a bioinspired, multi-segmented, compliant laminate sensor with embedded capacitive angle sensors. CITRAS is compact (\SI{73.7}{\times}\SI{15.6}{\times}\SI{2.1}{\milli\meter}), lightweight (\SI{491}{\milli\gram}), and low-power (\SI{32}{\milli\watt}), enabling seamless integration with miniature robotic platforms. The segmented compliant structure passively bends in response to environmental stimuli, achieving accurate hinge angle measurements with maximum errors of just \SI{0.79}{\degree} (quasistatic bending) and \SI{3.58}{\degree} (dynamic bending). Experimental evaluations demonstrate CITRAS' multifunctional tactile perception capabilities: predicting base-to-tip distances with \SI{7.75}{\percent} error, estimating environmental gap widths with \SI{6.73}{\percent} error, and distinguishing surface textures through differential sensor response. The future integration of this bioinspired tactile antenna in insect-scale robots addresses critical sensing gaps, promising enhanced autonomous exploration, obstacle avoidance, and environmental mapping in complex, confined environments.
\end{abstract}

\section{Move By Feel - Need for Tactile Perception}
\label{sec:intro}

Insect-scale robots (\SI{<2}{\centi\meter}, \SI{<2}{\gram};  \cite{doshi_effective_2019, jayaram_scaling_2020, singh2024buffalo, kim2024picotaur, liang_electrostatic_2021, johnson_millimobile_2023}) have significantly advanced in their capabilities in recent years \cite{de_croon_insect-inspired_2022}, 
now successfully navigating environments whose complexity and confinement approach those of their natural counterparts. For instance, drawing inspiration from the compliant exoskeletons of arthropods, recent miniature robots are now capable of adaptive morphological changes, enabling unprecedented locomotion in confined spaces \cite{jayaram_cockroaches_2016}. Notable examples include shape-morphing robots such as CLARI \cite{kabutz_design_2023} and its miniature variant mCLARI \cite{kabutz_mclari_2023}, capable of lateral body compression to navigate narrow horizontal gaps. Such small-scale robots offer new opportunities for robotics, including environmental monitoring \cite{aucone2023drone}, high-value asset inspection \cite{de_rivaz_inverted_2018}, search-and-rescue operations \cite{st2019toward}, and targeted healthcare delivery \cite{bozuyuk2024roadmap}. Despite these advances, reliable autonomous operation remains elusive due to severe size, weight, and power (SWAP) constraints, significantly limiting onboard sensing and perception capabilities.

\begin{figure}[t]
\centering
\includegraphics[width=0.9\linewidth]{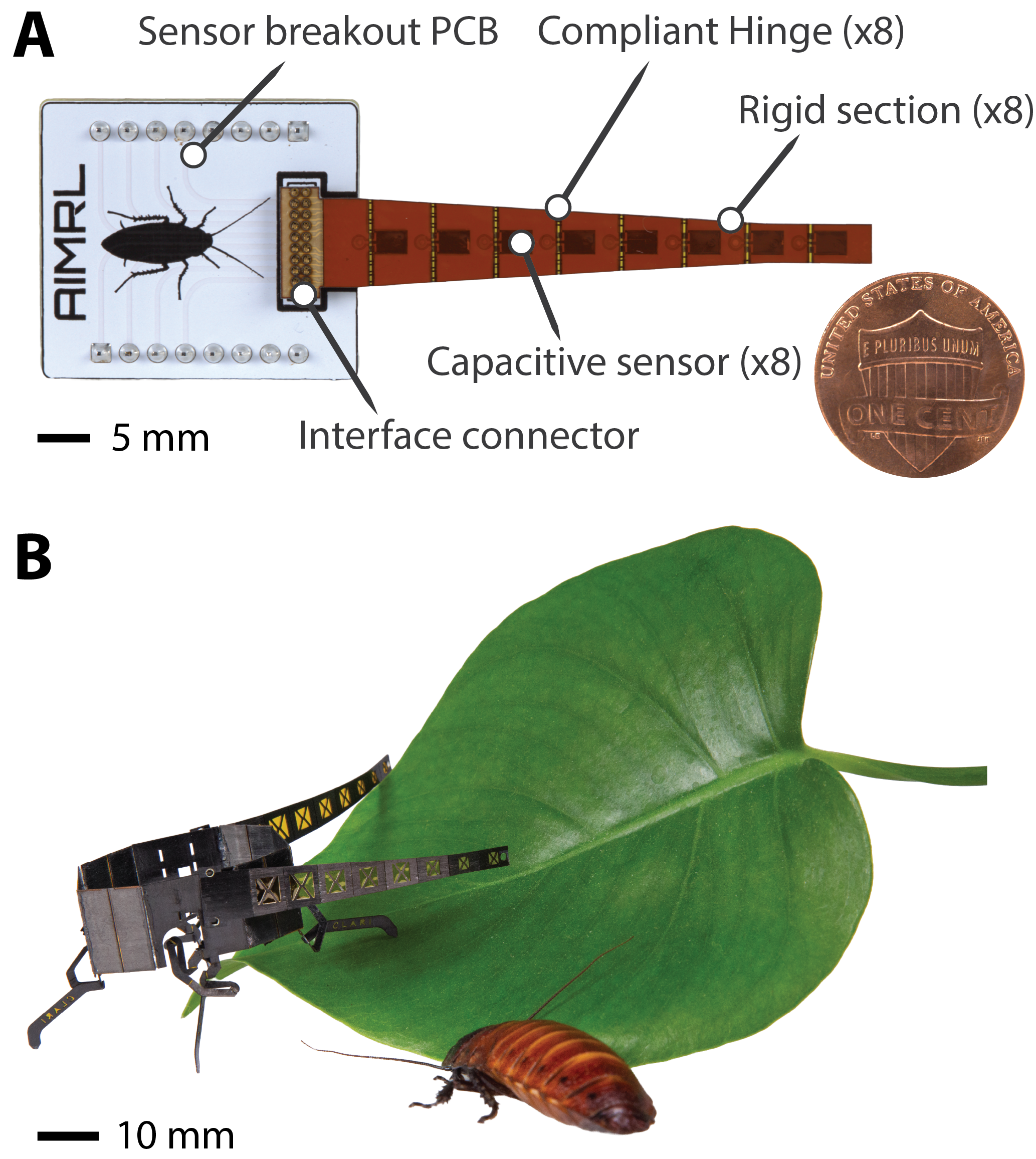}
\caption{Robotic antenna prototype overview and concept of operation \textbf{(A)} Final robotic antenna prototype assembly and scale reference. \textbf{(B)} Example of a robotic antenna integrated on mm-scale robot for future tactile exploration applications.}
\label{fig:overview}
\end{figure}

To achieve reliable autonomy in complex real-world environments, insect-scale robots require robust sensing capabilities explicitly tailored to their SWAP constraints. Vision-based systems have traditionally dominated robotic autonomy, proving highly effective in larger humanoid \cite{gao2022tactile}, dog \cite{miki2022learning} and cat-scale quadrupedal platforms \cite{margolis2022learning}. However, on the insect scale, vision-based sensing remains impractical due to substantial demands on power, weight, and onboard computational resources \cite{fuller2022gyroscope}. Moreover, visual sensors can quickly degrade in reliability under low-light, confined, or adverse environmental conditions commonly encountered in real-world scenarios \cite{zhang_perception_2023}. To address these limitations, larger robotic systems typically employ sensor fusion strategies that integrate multiple sensing modalities, such as LiDAR, optical depth sensors \cite{de_simone_one_2024}, and ground-contact sensors \cite{valsecchi_quadrupedal_2020}. Unfortunately, such traditional sensors are inherently bulky, energy-intensive, or both, making them unsuitable for direct downscaling to insect-sized robotic platforms \cite{st2019toward, bozuyuk2024roadmap}. Consequently, developing lightweight, low-power yet robust sensing solutions that can function effectively under these challenging constraints remains a critical research challenge.

Despite these challenges, recent advances have produced successful examples of miniature sensors specifically tailored to insect-scale platforms, including photodiodes for ambient light tracking \cite{johnson_millimobile_2023, fuller_estimating_2013}, miniature time-of-flight sensors for altitude estimation \cite{helbling_altitude_2018}, inertial measurement units for orientation control \cite{goldberg_power_2018}, biohybrid sensors enabling olfactory perception \cite{anderson_bio-hybrid_2020}, integrated actuator proprioception systems \cite{jayaram2018concomitant, kabutz_integrated_2025}, and visual-inertial attitude control \cite{fuller2022gyroscope, yu2025tinysense}. Although these sensing approaches meet strict SWAP constraints, their functionality remains limited for tasks such as close-range obstacle detection, tactile surface identification, and robust environmental mapping. For example, miniature single-axis LiDAR sensors, while capable of distance measurements up to \SI{\approx 2}{\meter}, frequently suffer performance degradation under bright sunlight conditions because of low optical power. Similarly, microelectromechanical ultrasonic rangefinders, currently available in compact packages (\SI{3.5}{\times}\SI{3.5}{\times}\SI{1.3}{\milli\meter}), often produce inaccurate or false readings when encountered with soft or perforated surfaces due to acoustic absorption effects. Thus, while non-contact sensing technologies continue to improve, they alone have yet to provide the reliable close-range environmental perception necessary for insect-scale robots to autonomously navigate cluttered and confined spaces \cite{st2019toward, bozuyuk2024roadmap}.

Biological tactile sensors, such as vertebrate whiskers \cite{hartmann2001active, grant2022can, adachi2022whiskers} and arthropod antennae \cite{mongeau_mechanical_2014, mongeau_sensory_2015}, provide animals with exceptionally rich spatiotemporal tactile information, allowing effective navigation and environmental perception. Inspired by these biological systems, numerous robotic tactile sensors have been developed, predominantly based on whisker-like architectures due to their conceptual simplicity. For instance, Kent et al. developed a vision-based whisker array capable of detecting contact and airflow stimuli using optical tracking \cite{kent_whisksight_2021}. Despite promising functionality, the system’s reliance on consistent illumination and computationally intensive image processing renders it impractical for untethered insect-scale applications. Similarly, Sullivan et al. demonstrated an active whisker-inspired tactile sensor employing a Hall-effect magnetic sensor integrated at the base of a motorized whisker, enabling effective texture and distance discrimination \cite{evans_effect_2013}. However, each sensor unit’s considerable mass (\SI{\approx 8}{\gram}) and high power demand (\SI{\approx 0.9}{\watt}) significantly exceed insect-scale SWAP constraints. Zhu et al. introduced a self-powered triboelectric bionic antenna capable of obstacle avoidance and material identification without external power, yet it was limited by a single sensing element and slow response times (\SI{350}{\milli\second}), restricting the robot’s speed and precision in tactile localization \cite{zhu_self-powered_2023}. Recent work exploring adaptive compliance in tactile sensing further emphasizes the critical role of mechanical tuning in sensor functionality \cite{solomon_robotic_2006, mulvey_haven_2024}. Demir et al. presented a multi-segmented bioinspired antenna with tunable compliance and Hall-effect joint-angle sensors \cite{demir_tunable_2010, demir2011tunable}, successfully demonstrating tactile surface mapping and wall-following capabilities \cite{lamperski_dynamical_2005}. However, the bulky sensing elements and the rigid structure impede effective miniaturization. More recently, Shahmiri et al. developed a flexible capacitive sensor array (ShArc) capable of multipoint curvature estimation, leveraging flex-circuit fabrication \cite{shahmiri_sharc_2020}. Although highly promising, ShArc’s performance was constrained by its low sampling rate (\SI{10}{\hertz}) and structural complexity, limiting dynamic capabilities and further miniaturization. Thus, despite notable advances, existing bioinspired tactile sensors still fall short of the specific requirements for insect-scale operation, particularly in terms of compactness, power efficiency, and bandwidth \cite{st2019toward, bozuyuk2024roadmap}.

To address the critical gap in insect-scale tactile sensing, we present CITRAS (Cockroach Inspired Tactile Robotic Antenna Sensor), a bioinspired, multi-segmented, compliant tactile antenna specifically aimed at increasing the situational awareness of insect-scale robotic platforms. Drawing direct inspiration from the antenna mechanics and morphology of the American cockroach (\textit{Periplaneta americana}), \cite{mongeau_mechanical_2014, mongeau_sensory_2015, meng2025mechanical}), CITRAS features a linearly decreasing stiffness profile, fabricated through a monolithic laminate manufacturing technique \cite{wood_microrobot_2008, whitney_pop-up_2011}. CITRAS comprises eight compliant flexural hinges, each integrated with high-resolution axial capacitive strain sensors capable of detecting femtofarad-level (\SI{\approx 1}{\femto\farad}) capacitance changes induced by hinge deflection. This highly sensitive instrumentation allows CITRAS to capture and quantify both static and dynamic mechanical stimuli effectively, despite its compact thickness of only \SI{2.11}{\milli\meter}. Like its biological counterpart, CITRAS passively adapts its shape to environmental contact, enabling robust tactile interactions including obstacle distance estimation, profile reconstruction, and surface texture discrimination. Crucially, CITRAS meets the stringent SWAP constraints inherent to insect-scale robots, significantly enhancing their autonomous tactile perception capabilities and paving the way for robust navigation in complex, confined environments.

\section{Insect-inspired Miniature Distributed Tactile Probe}

In this section, we present the design, theory of operation and fabrication of CITRAS, a bioinspired tactile antenna specifically engineered as a distributed mechanosensor array with integrated routing for signal transmission and measurement (Figure \ref{fig:design}). Using bioinspiration from the cockroach antenna morphology, CITRAS aims to replicate key biological advantages, including compliance, spatially distributed sensing, and multifunctional tactile perception. We specifically address the critical challenges associated with integrating robust high-resolution sensing into miniature laminate structures \cite{wood_microrobot_2008, whitney_pop-up_2011}, essential for scalability and practical deployment in insect-scale robotic platforms \cite{jayaram_scaling_2020, kabutz_mclari_2023}. Here, we describe our innovative approach to overcome these challenges, detailing both design considerations and manufacturing methods to enable precise tactile sensing within stringent SWAP constraints.

\subsection{Design and Sensor Modeling}
The core objective of the CITRAS design is to replicate the multifunctional tactile capabilities of cockroach antennas within the stringent SWAP constraints of insect-scale robots. To achieve these goals, we considered two critical aspects: (1) the antenna’s mechanical structure, including hinge segmentation and tapered stiffness, and (2) the choice and optimization of sensor transduction mechanisms. In the following, we outline our mechanical modeling approach, followed by a justification for choosing capacitive sensing and a detailed explanation of sensor theory and operation.

\subsubsection{Mechanical Model of Tapered Antenna Stiffness}

The mechanical design of the segmented CITRAS antenna draws inspiration from the flexural stiffness gradient observed in the antenna of the American cockroach, where flexural stiffness decreases exponentially from base to tip. The flexural stiffness gradient allows the cockroach antenna to passively conform to environmental features, simplifying tactile perception through morphological computation and enhanced spatial resolution \cite{mongeau_mechanical_2014, meng2025mechanical}.

To replicate this advantageous mechanical behavior within practical fabrication constraints, we adopted a linearly decreasing stiffness profile in CITRAS, explicitly tailoring the beam geometry and hinge dimensions along its length. This linear stiffness gradient significantly simplifies tactile sensing: segments closer to the antenna base provide structural support and efficiently transmit forces, while more compliant distal segments readily conform to environmental contours, thereby enhancing tactile sensitivity and spatial resolution near the antenna tip. Previous analytical and robophysical studies have characterized the mechanical effects of antenna tapering and its direct impact on tactile sensing performance \cite{mongeau_mechanical_2014, demir_tunable_2010}. Guided by these studies, CITRAS’s design parameters, hinge width linearly decreasing from \SI{8.0}{\milli\meter} at the base to \SI{3.62}{\milli\meter} at the tip, uniform thickness of \SI{25}{\micro\meter}, and hinge length of \SI{150}{\micro\meter}, were experimentally optimized to achieve the desired balance between stiffness, compliance, and tactile sensitivity. These design choices guided by insect antennae could enable robust and efficient tactile exploration in constrained and cluttered environments typical of insect-scale robotic operations.

\subsubsection{Capacitive sensing as preferred mechanosensor transduction method}
Tactile sensors, both biological and engineered, transduce mechanical interactions into measurable signals through various mechanisms, including capacitive displacement detection \cite{shahmiri_sharc_2020}, piezoelectric charges \cite{wang_tactile_2023}, Hall-effect sensing \cite{ren_robust_2025}, resistive strain gauges \cite{liu_bioinspired_2022}, and optical methods \cite{li_all-optical_2024}. Piezoelectric sensors, while highly sensitive to dynamic stimuli, typically exhibit poor low-frequency response, limiting their applicability for static or slow tactile interactions crucial for insect-scale navigation. Hall-effect sensors offer reliable angle measurements, but often necessitate bulky permanent magnets or ferrous-loaded composites, significantly exceeding the size and weight constraints inherent to miniature robots. Resistive strain sensors, though straightforward to implement, typically suffer from high thermal noise, low sensitivity, and limited operational lifespans due to fatigue-induced degradation, compromising reliability in prolonged field applications. Optical strain measurement systems, despite their high resolution, currently require bulky external components and sophisticated integration, making them impractical for miniaturized onboard deployment. In contrast, capacitive displacement sensing presents a highly advantageous solution for insect-scale applications due to its inherent characteristics of low power consumption, compact form factor, high linearity, high sensitivity, and absolute sensing capability, thus avoiding the need for continuous recalibration or complex incremental tracking \cite{fleming_review_2013}. These features make capacitive sensing the most suitable choice for integration within the strict SWAP constraints of CITRAS, providing robust and reliable tactile feedback necessary for autonomous navigation at small scales.

\begin{figure*}[ht!]
\centering
\includegraphics[width=1\textwidth]{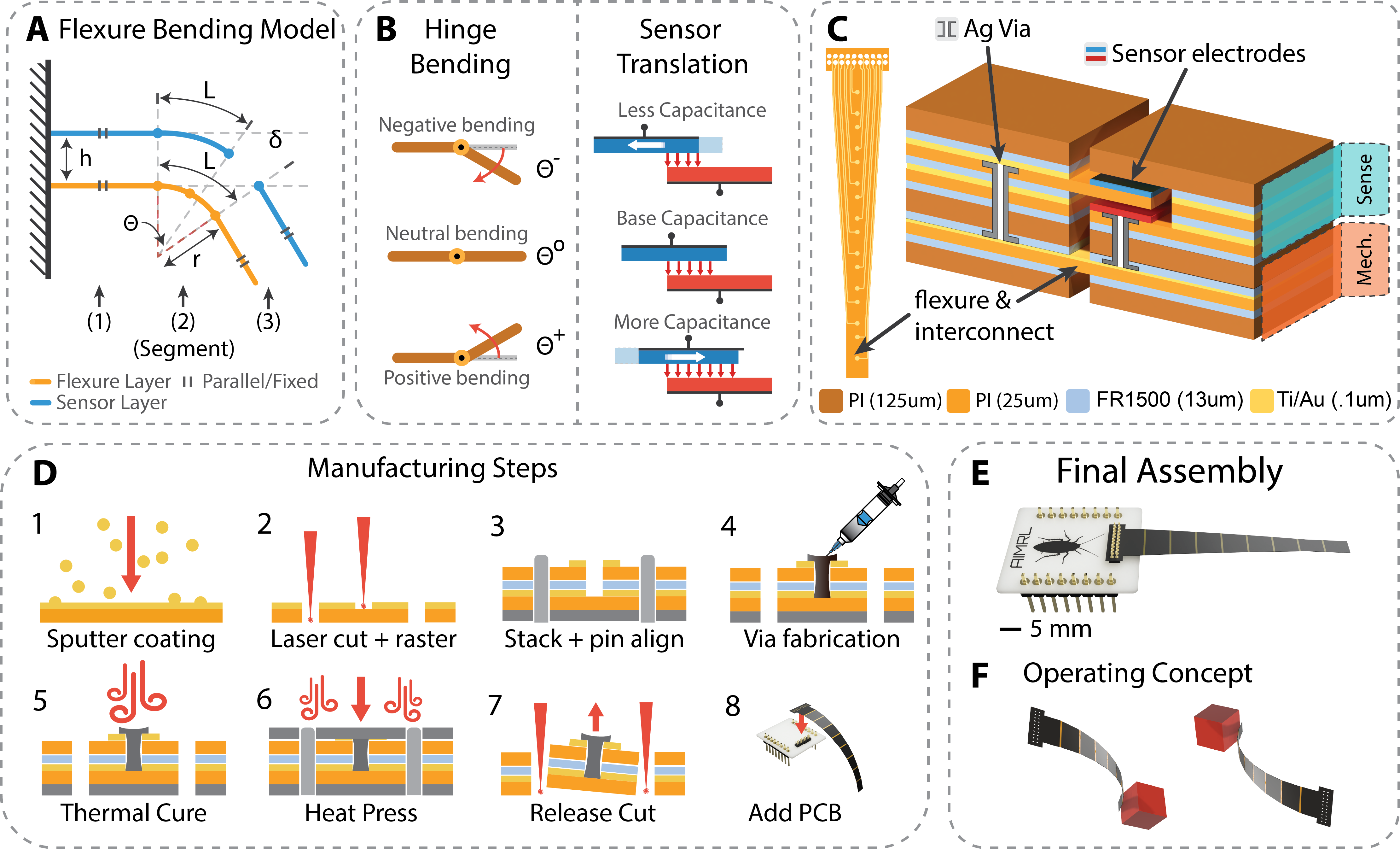}
\caption{Overview of sensor theory of operation, material stackup and  manufacturing methods.  \textbf{(A)} Hinge and sensing layer flexure mechanics and notation. \textbf{(B)} Simplified sensor concept of operation during positive and negative deflection. \textbf{(C)} Antenna material stackup of single hinge. \textbf{(D)} Manufacturing processes involved in antenna fabrication. \textbf{(E)} Render of complete antenna assembly with sensor breakout board.  \textbf{(F)} Demonstration of in-plane bending and environmental interaction of antenna. }
\label{fig:design}
\end{figure*}

\subsubsection{Sensor Theory of Operation}
\label{sensor_theory}


CITRAS integrates eight identical capacitive mechanosensors, each positioned at flexural hinges along its segmented beam structure, to measure the joint angle at its location. To illustrate the sensor operation and theoretical basis, we consider a single hinge mechanosensor unit, which fundamentally functions as a parallel-plate capacitor. Each mechanosensor consists of two conductive layers: a fixed electrode (the flexure layer in tan) and a sliding electrode (the sensor layer in cyan), vertically separated by a defined spacing $h$ (Figure \ref{fig:design}A).

As the hinge bends by an angle $\Delta\theta$ (starting from a neutral angle of 0), the sensor layer moves relative to the flexure layer, changing the overlap between these electrode plates (Figure \ref{fig:design}B). This overlap variation directly alters the mechanosensor's effective capacitance, $C$, according to the parallel-plate capacitor relationship:
\begin{equation}\label{eq:1}
C = \frac{\epsilon_0 \epsilon_r w l}{d}
\end{equation}
where $\epsilon_0$ represents the vaccuum permittivity, $\epsilon_r$ is the relative permittivity of the dielectric, $w$ is the electrode plate width, $l$ is the overlapping length and $d$ is the plate spacing.

The overlap length changes proportionally with the hinge bending angle, $\theta$:
\begin{equation}\label{eq:2}
l(\theta) = l_0 + \delta(\theta) = l_0 + h \Delta\theta
\end{equation}
where, $l_0$ is the initial overlap length (at zero deflection), and $\delta$ is the change in overlap resulting from hinge bending. Substituting this overlap relationship (Equation \ref{eq:2}) into the capacitance equation (Equation \ref{eq:3}) yields a relationship between capacitance ($C$) and bending angle:
\begin{equation}\label{eq:3}
C(\theta) = C_0 + \frac{\epsilon_0 \epsilon_r w\,h\,\Delta\theta}{d}
\end{equation}
where $C_0$ is the nominal capacitance at zero deflection. By performing an initial calibration (see the experimental section, Section \ref{sec:experimental_setup}), $C_0$ is determined and typically subtracted off, producing a linear relationship between capacitance change ($\Delta C = C-C_0$) and bending angle ($\theta$) allowing us to determine joint angle change by measuring capacitance change as: 
\begin{equation}\label{eq:4}
\Delta\theta = \frac{\Delta C d}{\epsilon_0 \epsilon_r w h}
\end{equation}

By integrating eight of these mechanosensors along the segmented antenna, CITRAS achieves distributed tactile sensing, enabling precise reconstruction of antenna shape profiles and accurate estimation of contact angles and environmental distances.

\subsection{Fabrication}

CITRAS is fabricated using a precision laminate manufacturing approach inspired by methods previously developed for insect-scale robotic devices \cite{wood_microrobot_2008, whitney_pop-up_2011}. This method enables the precise integration of multiple functional materials into compact, lightweight structures suitable for insect-scale platforms. The resulting laminate structure mimics the physical characteristics of biological antennae, highlighting their potential as near-scale robophysical models \cite{aguilar2016review} capable of high-fidelity hypothesis testing without the need for dynamic scaling.

\subsubsection{Laminate Functional Architecture}

Each CITRAS antenna consists of 14 individual material layers, vertically stacked and bonded into a cohesive laminate structure. The detailed laminate stackup for a representative hinge segment is depicted in Figure \ref{fig:design}C, illustrating individual material layers and their functional grouping. The laminate structure is functionally organized into two primary categories: (1) mechanical and electrical interconnection layers (layers 1–6) and (2) capacitive angular sensing layers (layers 7–14). The mechanical layers form the compliant flexural hinges that serially link nine rigid antennal segments, constraining bending to planar (two-degree-of-freedom) motion at eight specific locations along the antenna’s length.

\subsubsection{Process and Assembly}
Key fabrication innovations in CITRAS include the reliable creation of vertical electrical interconnects (vias) for robust signal transmission from individual mechanosensor units to the antenna base, as well as precise electromechanical alignment across multilayer laminate structures to ensure optimal sensor performance and structural integrity.
The major steps of the CITRAS fabrication pipeline are illustrated in Figure \ref{fig:design}D and described in detail below. 

We first selected specific materials and designed individual layers based on mechanical, electrical, and sensing requirements. For inter-layer bonding, we chose an acrylic adhesive film (Pyralux FR1500, DuPont, thickness: \SI{12.7}{\micro\meter}) due to its low mass, high peel strength, and minimal adhesive flow during curing. For rigid structural layers, we selected polyimide (Kapton, DuPont, thickness: \SI{127}{\micro\meter}) due to its adequate mechanical strength and rapid laser processing time. Additionally, a thinner polyimide film (thickness: \SI{25}{\micro\meter}) was specifically chosen for the compliant mechanical flexure layers, due to its favorable mechanical properties and its ability to withstand repeated mechanical bending cycles.
For sensing and electrical interconnection layers, we employed commercially available gold-coated polyimide substrates (Stanford Advanced Materials; polyimide thickness: \SI{25}{\micro\meter}, initial gold thickness: \SI{25}{\nano\meter}). To enhance electrical conductivity, an additional \SI{75}{\nano\meter} gold layer was sputtered onto these substrates, achieving a total conductor thickness of \SI{100}{\nano\meter} (Figure \ref{fig:design}D1).

To create the vertical interconnect sensor sublaminate, individual layers were selectively patterned (surface rastered or cut through, as appropriate), achieving precise geometric shapes and alignment features (Figure \ref{fig:design}D2). Specifically, the conductive gold layer was raster etched from the polyimide substrates, forming electrode patterns, conductive traces, and vias. Adhesive and rigid structural layers were similarly laser-machined to obtain the required geometries. To ensure accurate stacking, identical alignment pin holes were cut in all layers during laser processing (Figure \ref{fig:design}D3).
After fabricating individual layers, we performed a precise, layer-by-layer stacking procedure using a custom pin alignment jig (Figure \ref{fig:design}D3). The stacked layers were partially cured in a controlled heat press (Master Press; \SI{12}{\minute} at \SI{30}{PSI} and \SI{300}{\degree\fahrenheit}) to establish sufficient interlayer adhesion, preventing conductive ink leakage between layers during subsequent via formation (Figure \ref{fig:design}D4). Reliable vertical electrical interconnections (one via per mechanosensor unit) were established by dispensing semi-sintering silver conductive ink (ACI Materials FS0142) using a pneumatic dispenser (SRA, APPLSRA205-220V). After via filling, the partially cured laminate underwent an additional curing step in a thermal oven (\SI{15}{\minute} at \SI{150}{\degree\celsius}) to fully solidify the conductive ink and establish robust vertical connections (Figure \ref{fig:design}D5). This sequential stacking, partial curing, and via formation procedure was then repeated for the top mechanosensor electrode layers.
Protective polyimide layers were subsequently added to provide mechanical protection and limit unwanted vertical electrode movement during hinge deflection. To prevent restriction of sliding electrode movement, custom cutouts in the heat press tooling were created above each mechanosensor location, reducing applied pressure in these sensitive regions.

The mechanical sublaminate was separately manufactured following a similar laminate procedure, then bonded with the sensor sublaminate in a final heat press cycle (\SI{90}{\minute} at \SI{30}{PSI} and \SI{325}{\degree\fahrenheit}, Figure \ref{fig:design}D6). A final precision laser cut was performed to release the integrated electromechanical structure from the surrounding material (Figure \ref{fig:design}D7). Released antenna structures were press-fit onto custom-designed, 20-pin, \SI{1}{\milli\meter}-pitch connectors mounted on breakout printed circuit boards (PCBs). Electrical connectivity between antenna mechanosensor units and PCBs was established by extruding conductive ink over each connector pin and curing, forming robust signal transmission pathways (Figure \ref{fig:design}D8).

The entire fabrication process, from initial material patterning to the completed electromechanical antenna, requires approximately six hours, yielding three fully operational CITRAS antennas per fabrication cycle. This systematic approach ensures robust, consistent, and scalable production of antennas specifically suited for insect-scale tactile sensing and robotic applications.

\begin{figure*}[ht!]
\centering
\includegraphics[width=1\textwidth]{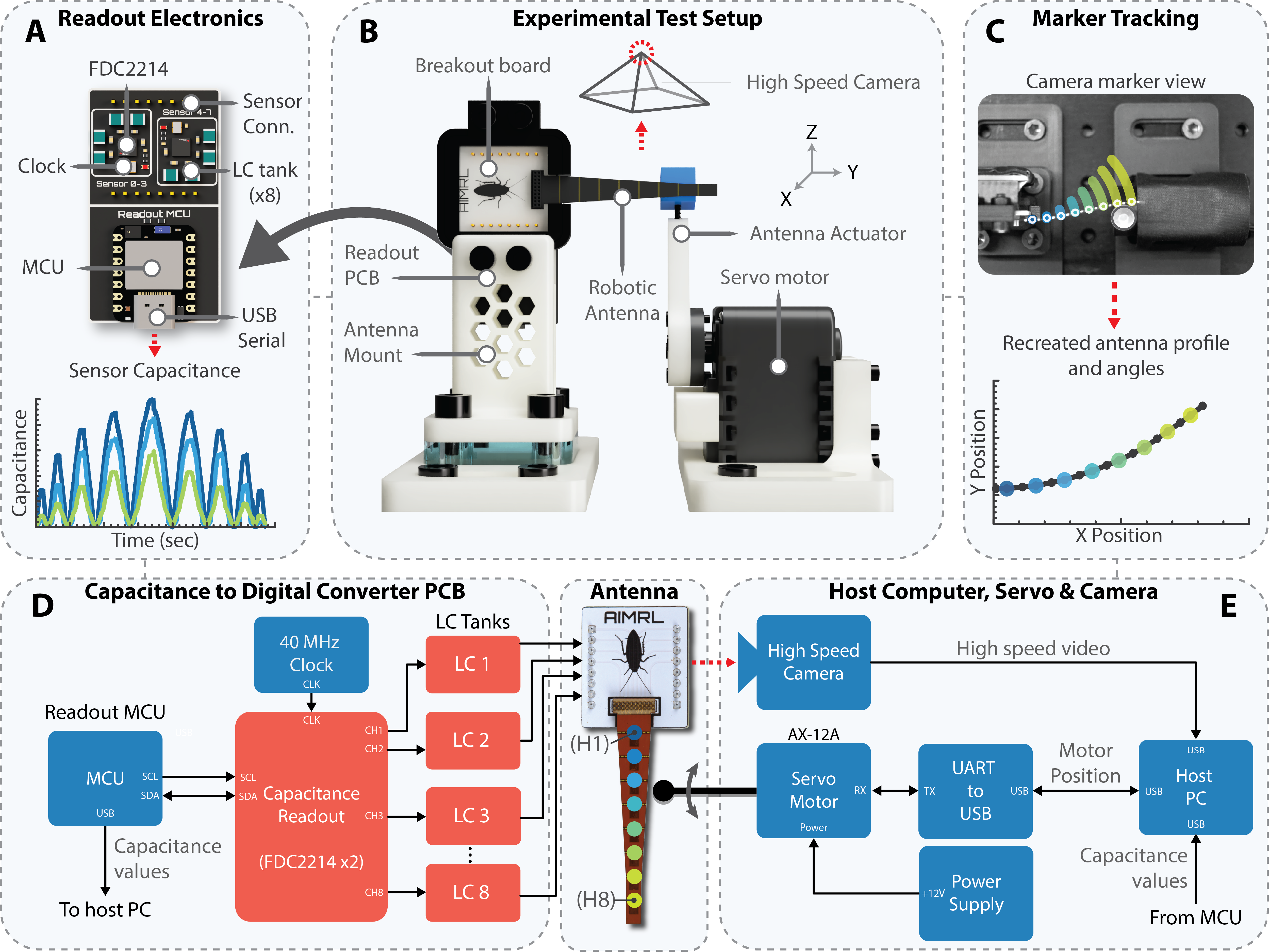}
\caption{Experimental setup overview \textbf{(A)} Eight channel capacitive to digital converter (CDC) PCB records sensor capacitance and transmits data over serial connection to host PC. \textbf{(B)} Schematic view of experimental setup configured to collect data for quasistatic and dynamic results. \textbf{(C)} View of antenna and tracking markers from speed camera perspective. DLTdv was used to extract marker X-Y location from video. \textbf{(D)} Detailed block diagram of experimental setup subsystems and interconnections.} 
\label{fig:experimental_setup}
\end{figure*}

\subsubsection{Final Prototype}

Based on the theoretical model presented in Section \ref{sensor_theory} and the iterative experimental validation, we optimized the physical parameters of CITRAS to achieve a desirable balance between mechanical stiffness, compliance, sensitivity and ease of fabrication. The finalized dimensions explicitly chosen for the antenna are: hinge width linearly tapering from \SI{8.00}{\milli\meter} at the antenna base to \SI{3.62}{\milli\meter} at the distal tip, uniform hinge thickness of \SI{25}{\micro\meter}, and consistent hinge length of \SI{150}{\micro\meter}.
With these parameters, each mechanosensor unit is predicted to exhibit an angular sensitivity of approximately \SI{7.14}{\femto\farad/\degree}. Considering the inherent noise floor of our capacitive measurement circuitry (\SI{\approx 0.3}{\femto\farad}), each sensor can theoretically resolve angular deflections as small as \SI{0.04}{\degree/count}.
The final assembled CITRAS antenna, including integrated flex-PCB readout electronics, measures approximately \SI{73.7}{\milli\meter} in total length, \SI{15.6}{\milli\meter} in width and \SI{2.11}{\milli\meter} in thickness. Its total mass is \SI{491}{\milli\gram}, and its operational power consumption is approximately \SI{32}{\milli\watt}, ideal for SWAP-limited insect-scale platform integration.

\section{Distributed Tactile Sensing Performance} 
\label{experimental_results}

This section describes the experimental setup, methodologies, and procedures employed to evaluate the tactile sensing performance of CITRAS (Section \ref{sec:experimental_setup}. We systematically characterize the mechanosensors' performance under quasistatic (Section \ref{sec:quasistatic_characterization}) and dynamic bending conditions (Section \ref{sec:dynamic-characterization}), establishing their accuracy, resolution, and dynamic response capabilities. Finally, we examine the angular sensing limitations arising from mechanosensor saturation during large deflections in Section \label{large_angle_bending}. 

\subsection{Experimental Setup} 
\label{sec:experimental_setup}

The experimental setup used to characterize the mechanical and sensory characteristics of the CITRAS antenna is depicted in Figure \ref{fig:experimental_setup}. The setup includes custom mechanical fixtures for antenna mounting, specialized readout circuitry to measure individual mechanosensors, and a servo motor (Robotis, Dynamixel AX-12A) to apply controlled repeatable deflections along the antenna. All components were rigidly mounted on an aluminum breadboard (MB1503F/M, Thorlabs) to minimize unwanted movements. The setup was designed to be reconfigurable for different experiments, with specific modifications described in the relevant sections.

The CITRAS antenna was connected through a 20-pin connector at its base to a custom breakout printed circuit board (PCB) that interfaced directly with measurement electronics (Figure \ref{fig:experimental_setup}A). This PCB was bolted onto custom-designed 3D-printed fixtures, constraining the antenna’s motion to a two-dimensional plane (X/Y plane), as shown in Figure \ref{fig:experimental_setup}B. Changes in capacitance of each mechanosensor unit, at the femtofarad level, were measured using two four-channel capacitance-to-digital converter (CDC) chips (FDC2214, Texas Instruments). Capacitive measurements were sampled using an nRF52840 microcontroller module (XIAO BLE Sense, Seeed Studio) via a \SI{400}{\kilo\hertz} I²C communication bus and transmitted to a host computer through a high-speed USB serial interface, achieving an average sampling rate of approximately \SI{80}{\hertz} per mechanosensor.

Controlled antenna deflections were achieved by programming the servo motor equipped with spherical end-effector (diameter: \SI{10}{\milli\meter}) to specific trajectories and thus making contact with the antenna at the desired location and speed. The servo was powered by a \SI{12}{\volt} DC supply, with serial communication managed by a USB-to-serial converter (Robotis, U2D2). To obtain precise ground-truth hinge angles, a high-speed video camera mounted directly above the experimental setup captured movements of white markers painted at discrete locations along the antenna’s profile. Marker positions in recorded videos were tracked using the digitizing software DLTdv and processed using custom MATLAB scripts to reconstruct the antenna's profile and accurately determine the hinge angles.

Due to minor manufacturing variations, the true vertical spacing between the sensor electrodes may vary slightly across individual mechanosensor units and different antenna samples. Thus, each mechanosensor requires an individual calibration prior to testing, to ensure accurate angle determination and optimal sensor performance. 

\begin{figure}[!ht]
\vspace{-4mm}
\centering
\includegraphics[width=0.4\textwidth]{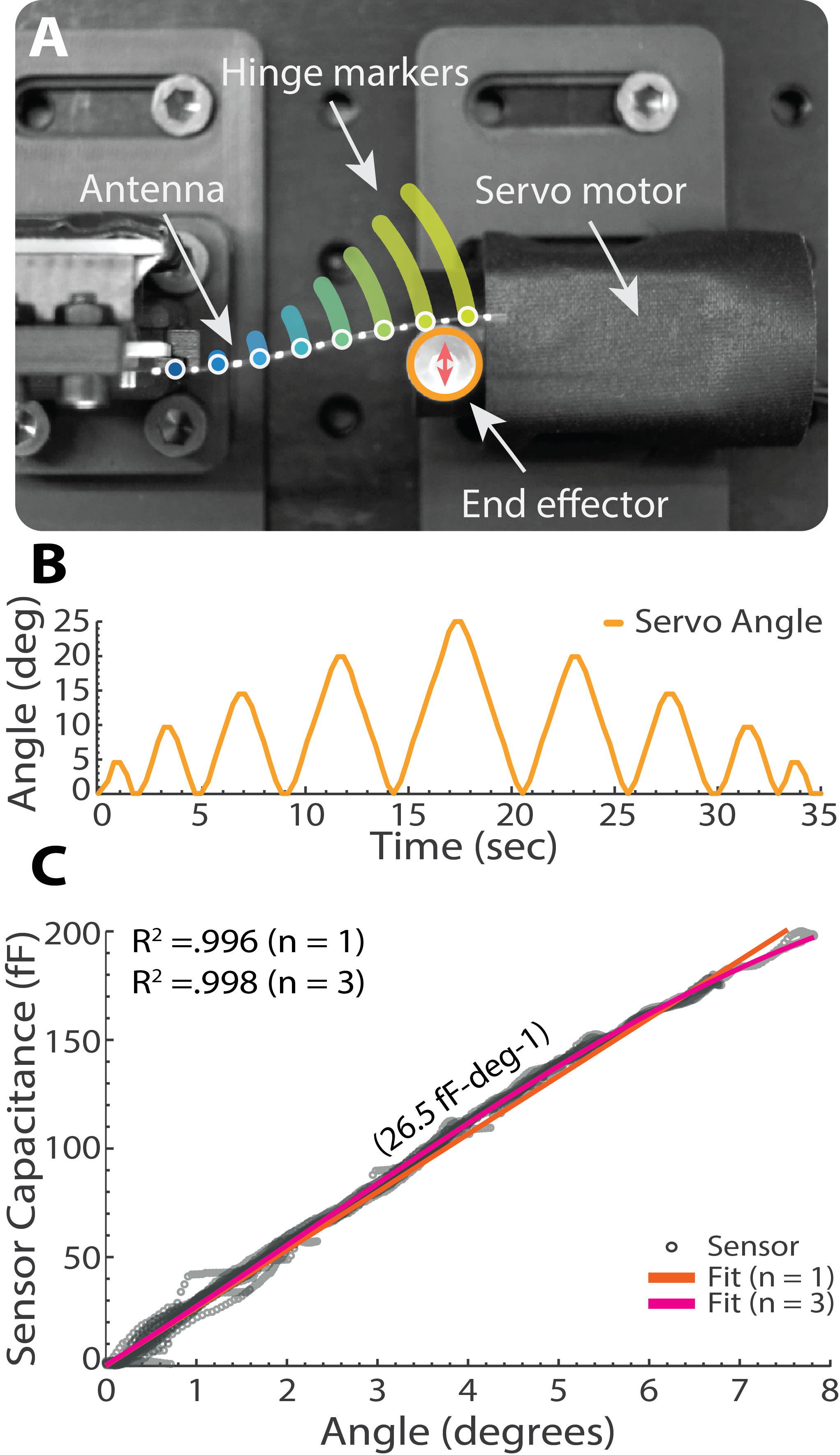}
\caption{Quasistatic sensor characterization experimental setup and results. \textbf{(A)} Quasistatic experimental setup shown from antenna tracking camera perspective. \textbf{(B)} Servo motor angle versus time profile during characterization experiment. \textbf{(C)} Capacitance change versus hinge angle data plotted for mechanosensor at the hinge closest to the base (H1). 
}
\label{fig:quasi-static-1}
\end{figure}

\subsection{Quasistatic Characterization} 
\label{sec:quasistatic_characterization}

We first characterized the quasistatic performance of the mechanosensor units to quantify their accuracy and repeatability under controlled bending conditions. A representative experiment is shown in Figure \ref{fig:quasi-static-1}A, where a servo motor-driven end-effector applied precise deflections to the antenna, rigidly mounted at the base, at defined locations along its length. The servo rotation followed a periodic triangular profile with a peak deflection of approximately \SI{25}{\degree}, as illustrated in Figure \ref{fig:quasi-static-1}B. Quasistatic conditions were ensured by applying a low motor rotational speed of \SI{0.12}{\radian/\second}, corresponding to a linear speed of \SI{4.6}{\milli\meter/\second} at the contact point. Mechanosensors H7 and H8 were excluded from the subsequent analyses, as controlled bending at these distal segments was unreliable due to experimental limitations (e.g., end-effector contact near the tip causes excessive slipping).

Figure \ref{fig:quasi-static-1}C shows representative capacitance changes for hinge H1 as a function of angular deflection, yielding a linear sensitivity (over small angles) of approximately \SI{26.5}{\femto\farad/\degree}. This observed sensitivity significantly exceeds the theoretical prediction (\SI{7.14}{\femto\farad/\degree}, Equation \ref{eq:4}), likely due to manufacturing imperfections and unintended variations in electrode spacing during bending. To systematically relate capacitance changes to hinge angles, third-order polynomial fits provided the best relationships, with coefficients of determination ($R^2$) consistently exceeding 0.99 across all hinges in the operational range.

\begin{figure}[!hb]
\centering
\includegraphics[width=0.45\textwidth]{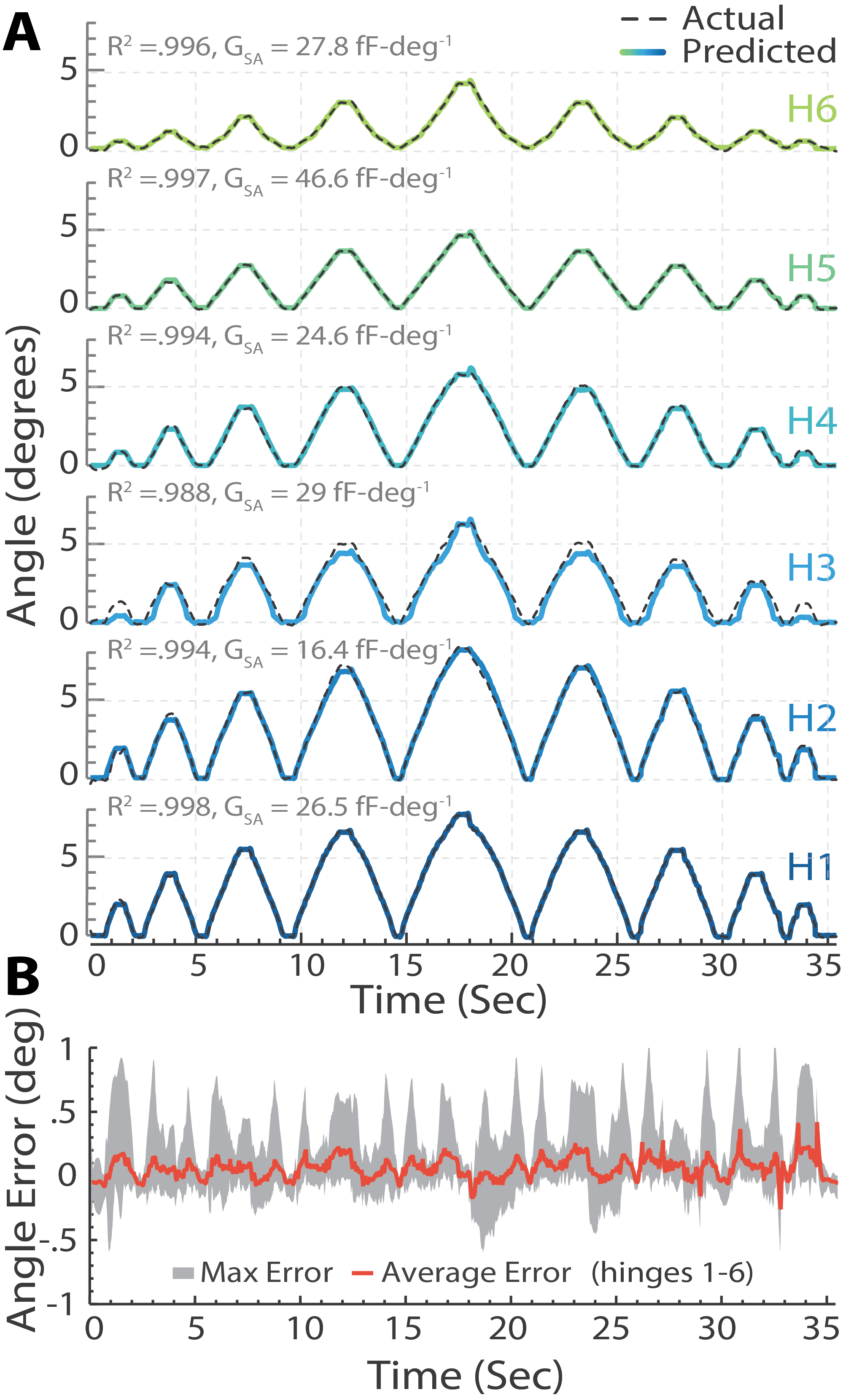}
\caption{Quasistatic sensor based angle prediction performance results. \textbf{(A)} Comparison of predicted and ground truth angles for actuated hinges (1-6).   \textbf{(B)} Maximum and average error plots across actuated hinges (1-6). A maximum error of $0.795\ degrees$ is measured on hinge $3$.}
\label{fig:quasi-static-2}
\end{figure}

Using these experimentally derived third-order transfer functions, hinge angles were predicted from raw mechanosensor capacitance data collected during small-angle deflections. Figure \ref{fig:quasi-static-2}A compares predicted hinge angles (solid lines) with ground truth data (dashed lines) for mechanosensors H1–H6, demonstrating strong agreement. To quantify performance, Figure \ref{fig:quasi-static-2}B presents hinge angle prediction errors as average and maximum error curves across hinges H1–H6. The average angle prediction error was minimal (\SI{0.056}{}$\pm$\SI{0.079}{\degree}; mean$\pm$1 s.d.), and maximum errors were similarly low (\SI{0.25}{}$\pm$\SI{0.28}{\degree}). These results confirm that mechanosensors accurately and reliably capture hinge angular states during quasistatic bending. Consequently, precise reconstruction of the antenna’s shape profile is achievable, facilitating accurate perception of environmental geometry and enabling robust tactile exploration and intelligent decision making in future insect-scale robotic platforms.

\begin{figure*}[!ht]
\centering
\includegraphics[width=0.95\textwidth]{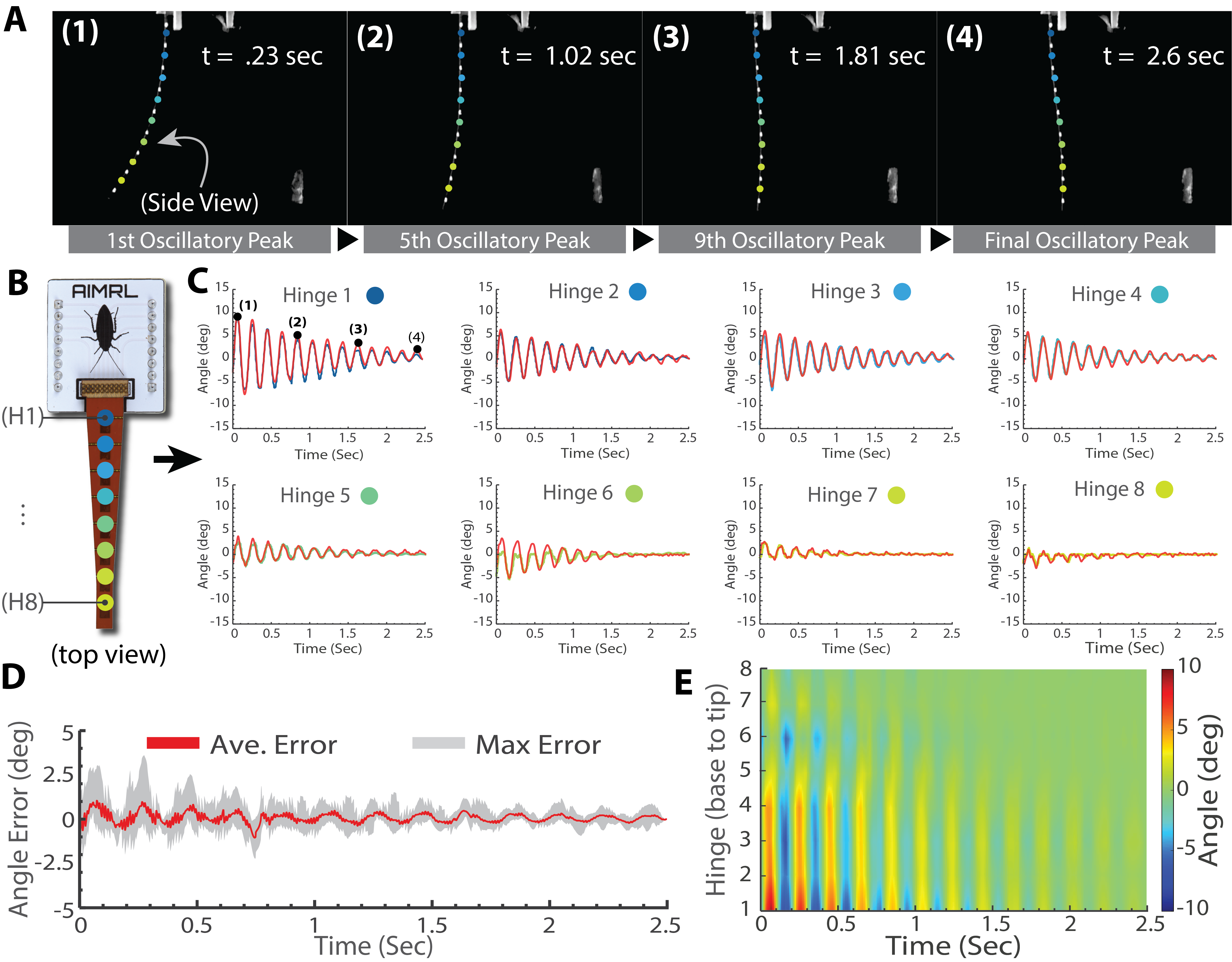}
\caption{Sensor response during dynamic bending  \textbf{(A)} High speed optical image sequence depicting antenna motion during dynamic bending. \textbf{(B)} Antenna hinge numbering legend from base to tip. \textbf{(C)} Comparison of sensor based prediction and ground truth hinge angle ($1-8$). \textbf{(D)} Angular error between actual and predicted hinge angles. Average error (red) and maximum error (gray) across all hinges.  \textbf{(E)} Spatiotemporal "Tactile Image" representation of antenna angular response over space and time. Unique tactile images are generated based on object type, contact location and speed.}
\label{fig:dynamic-characterization}
\end{figure*}

Finally, the above experiments reveal that calibration procedures are critical to compensate for manufacturing variations and therefore, each mechanosensor unit was individually calibrated prior to experimental evaluation to ensure accurate angle determination and optimize sensor performance.

\subsection{Dynamic Characterization} 
\label{sec:dynamic-characterization}

Having established mechanosensor performance during quasistatic conditions, we next characterized both the mechanical dynamics of the antenna and the sensor response to rapid deflections, conditions representative of realistic tactile navigation and interaction scenarios. The experimental setup for dynamic characterization was identical to the quasistatic tests described above. In these experiments, the servo motor displaced the antenna tip by approximately \SI{56}{\degree}, corresponding to a horizontal displacement of \SI{39.2}{\milli\meter} at H8. Upon reaching this displacement, contact was released, allowing the antenna to freely oscillate until coming to rest. 

A representative time sequence of this dynamic antenna motion is illustrated in Figure \ref{fig:dynamic-characterization}A. Video tracking data indicated second-order exponentially decaying sinusoidal responses for each hinge angle across the antenna (Figure \ref{fig:dynamic-characterization}B-C). 
We found a strong agreement with sensor-predicted hinge angles and ground truth measurements with an average error of \SI{0.10}{}$\pm$\SI{0.27}{\degree} indicating effective dynamic performance quantification with our approach.

Using standard system identification methods, we measured a natural frequency 
of \SI{31.71}{}$\pm$\SI{0.29}{\radian/\second} and  a damping ratio 
of \SI{0.035}{}$\pm$\SI{0.012}{}, averaged across all mechanosensor hinges.
Notably, the measured damping ratio is approximately an order of magnitude lower than values reported for the biological antenna of the American cockroach (\SI{\approx 0.3}{}; \cite{mongeau_mechanical_2014}). This difference arises primarily from the low viscoelasticity of the polyimide and low friction of hinge flexures compared to biological antenna tissues and structures. Such low damping prolongs the antenna's settling time after rapid deflection, potentially limiting the speed at which closed-loop tactile navigation could reliably occur. However, this damping characteristic can be coarsely tuned during fabrication: increasing bonding pressure applied in the heat press step enhances friction between sliding electrode layers, allowing practical adjustments to achieve desired damping levels. More effective future methods could be integrating highly viscoelastic elastomers \cite{lakes2009viscoelastic}, tunable smart materials \cite{hedrick2024femtosecond, mcclintock2021fabrication},  fluid-filled structural elements \cite{kellaris2021spider} and actively controlled mechanisms \cite{diller2016lightweight, hinchet2020high} into the fabrication of bioinspired antennae. 

To visualize and exploit the spatial-temporal patterns captured by the antenna's distributed mechanosensors during dynamic deflections, we invoke the concept of a \textit{tactile image} \cite{bach1969vision}. This representation plots the hinge deflection angles as a function of both spatial position along the antenna (hinge number, Y-axis) and time (X-axis), analogous to a waterfall plot. As seen in Figure \ref{fig:dynamic-characterization}E, the tactile image visually reveals unique and rich spatiotemporal patterns generated by interactions between the antenna and its environment, suggesting its potential utility for classification tasks using established computer vision techniques \cite{lu2007survey, rawat2017deep}. Such methods could enable robust tactile recognition, differentiating object geometry, contact location, texture, or even identifying specific interaction patterns, which are essential capabilities \cite{kent_whisksight_2021,mulvey_haven_2024, pearson2011biomimetic,yu_bioinspired_2022} for insect-scale autonomous navigation.

\subsection{Large angle deflection and sensing limits} \label{sec:large_angle_bending}

\begin{figure}[!b]
\centering
\includegraphics[width=0.78\linewidth]{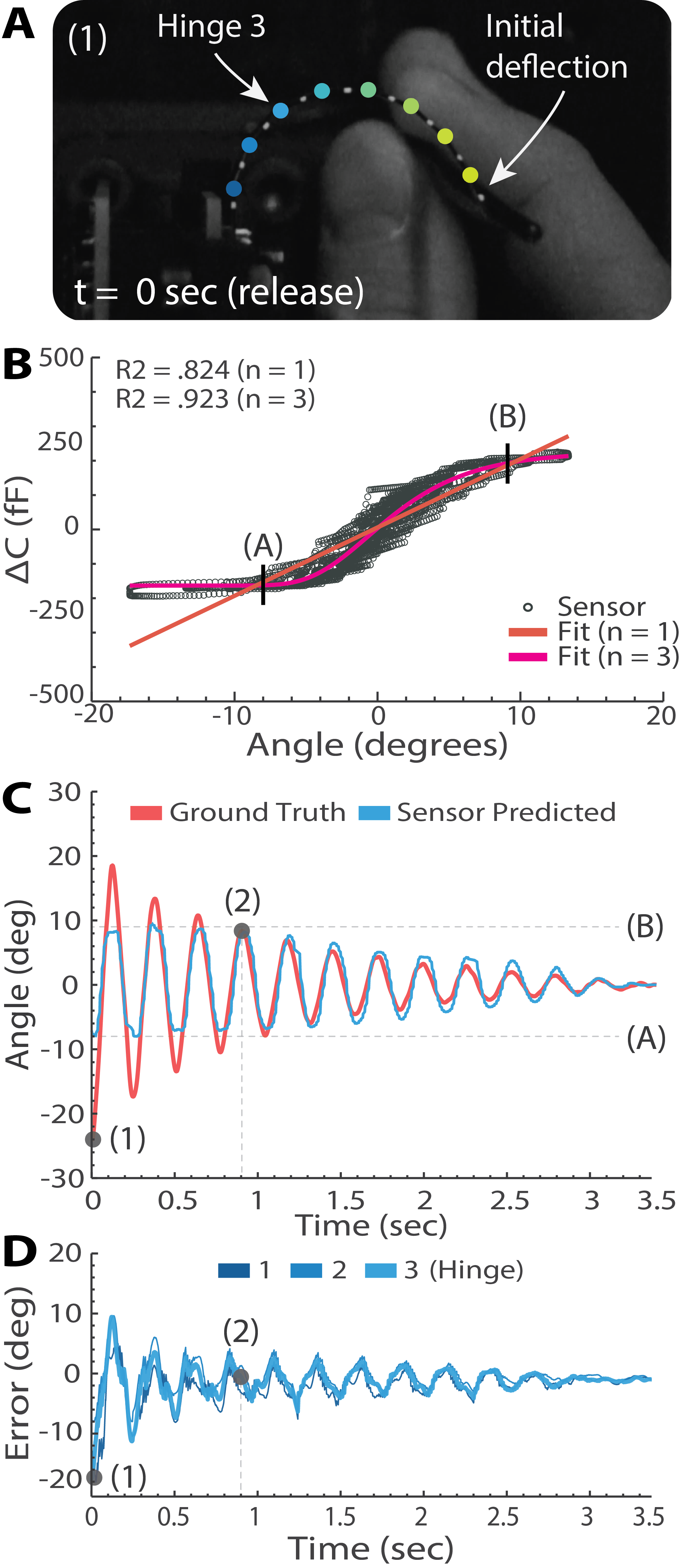}
\vspace{-3mm}
\caption{Sensor angle limit characterization during large bending. \textbf{(A)} Large antenna deflection at moment of release. \textbf{(B)} Sensor capacitive change versus angle response versus hinge angle curve. Estimated sensor linear limits indicated by $A\ \&\ B$. \textbf{(C)} Sensor predicted vs. ground truth hinge angle. Marker $2$ indicates estimated time when linear tracking is regained. \textbf{(D)} Absolute angular error versus time for hinges $1-3$.}
\label{fig:large-angle-deflection}
\end{figure}

To identify the operational limits of mechanosensor units under large-angle deflections, we conducted experiments involving substantial tip deflection. Specifically, the servo motor displaced the antenna tip by approximately \SI{88}{\degree} relative to the base hinge (H1), a deflection significantly beyond typical operational ranges (Figure \ref{fig:large-angle-deflection}A). Due to the mechanical design, hinges H1 through H3 experienced the largest initial angular deflections, measuring \SI{24.1}{\degree}, \SI{26.4}{\degree}, and \SI{23.7}{\degree}, respectively. Figure \ref{fig:large-angle-deflection}B illustrates a representative mechanosensor (H3) capacitance response curve as a function of hinge angle, showing measurable responses for small deflections within approximately \SI{\pm10}{\degree} and saturation beyond this range. Deflections at hinges H4 and distal segments remained within the linear saturation limits.

The impact of mechanosensor saturation was evident when predicting hinge angles from mechanosensor capacitance during dynamic testing (Figure \ref{fig:large-angle-deflection}C). Prior to \(t=\SI{1}{\second}\), a substantial discrepancy was observed between predicted (sensor-derived) and ground-truth hinge angles due to sensor saturation. Quantitative analysis of prediction errors for hinges H1–H3 (Figure \ref{fig:large-angle-deflection}D) revealed a temporary loss of accurate angle tracking during large deflections. However, once deflections returned within approximately \SI{\pm10}{\degree}, accurate angle predictions immediately resumed, indicating that saturation does not permanently degrade sensor accuracy or calibration.

Although mechanosensor saturation imposes practical limitations on accurate sensing for objects with large curvature or extreme proximity to environmental obstacles, these constraints can be mitigated in future designs. Specifically, mechanical modifications such as increased electrode overlap and extended axial electrode travel could extend the mechanosensors' operational dynamic range, improving linearity and overall performance.

\section{Applications}
\label{sec:applications}

The previously detailed characterization experiments demonstrate that CITRAS provides accurate and reliable tactile sensing capabilities. To further illustrate its potential for real-world robotic tasks, we evaluated CITRAS in several realistic tactile sensing scenarios relevant to confined space navigation with insect-scale robotic platforms. First, we investigated the sensor's ability to accurately predict body-to-wall distances through tactile interactions with periodic surfaces (Section \ref{sec:btwd}). Next, we examined the sensor's performance in estimating environmental gap widths, a capability inspired explicitly by biological antenna behaviors crucial for navigating confined spaces (Section \ref{sec:gap-width}). Finally, we demonstrate the potential of CITRAS to discriminate between surface textures, such as differentiating smooth from rough (Section \ref{sec:texture-discrimination}). Together, these practical experiments validate CITRAS's promising applicability to real-world tactile exploration tasks.

\subsection{Predicting Body-to-wall Distance through Touch}
\label{sec:btwd}

\begin{figure}[!b]
    \vspace{-3mm}
    \centering
    \includegraphics[width=1\linewidth]{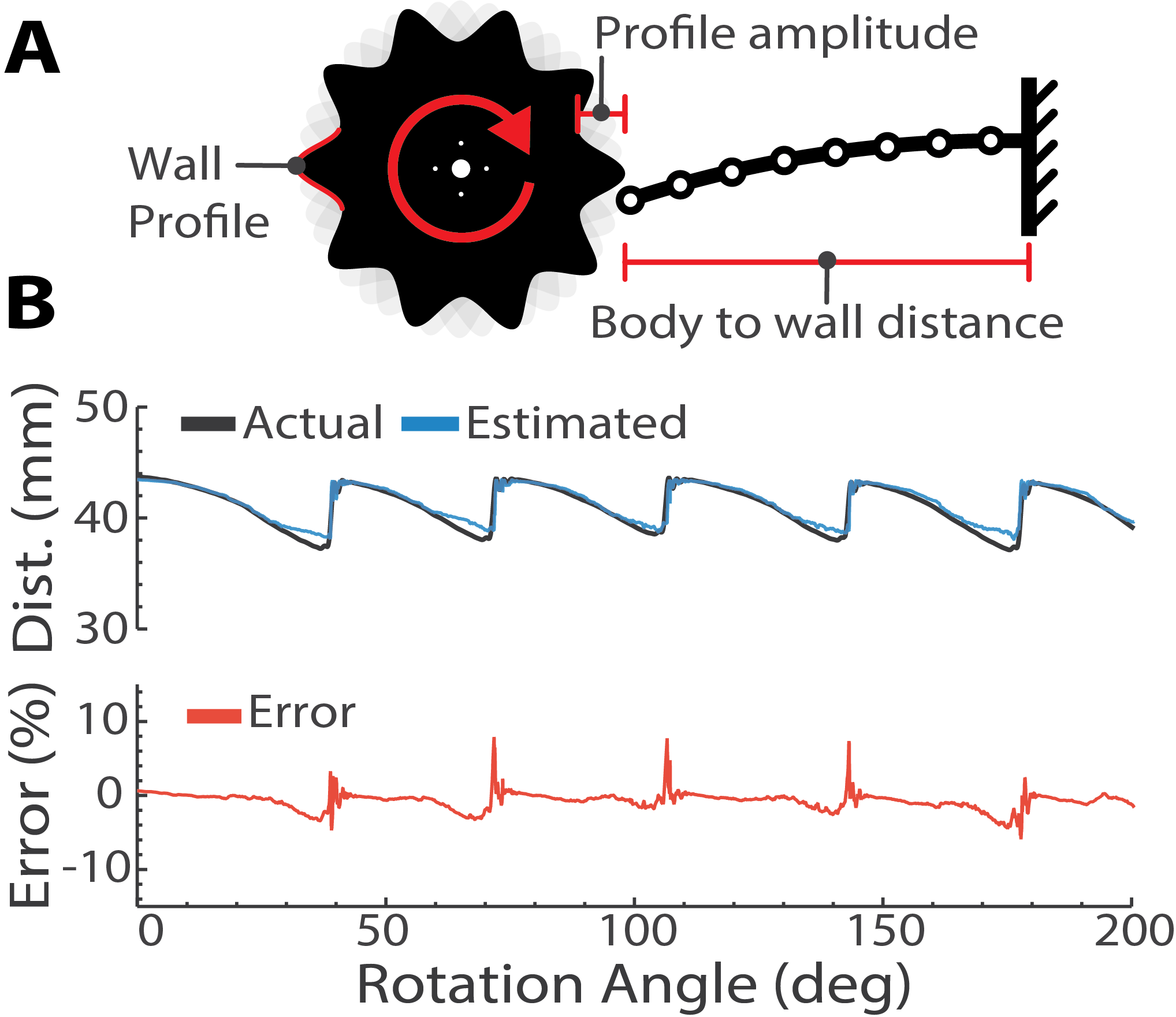}
    \caption{Prediction of body to wall distance (BTWD) with sensorized antenna \textbf{(A)} Schematic diagram of experimental setup with measurement definitions. \textbf{(B)} Comparison of actual and sensor based estimate of body to wall distance (top), and resulting percent error (bottom). For sinusoidal profile, predicted BTWD closely matches tracking marker based ground truth results (top), with a maximum percent error of $7.75\%$ (bottom).}
    \label{fig:btwd}
\end{figure}

Accurate estimation of distance from a robot’s body to surrounding environmental features is critical for robust tactile navigation tasks such as obstacle avoidance, exploration, and wall-following \cite{jusuk_lee_templates_2008}. To demonstrate CITRAS’s suitability for such tasks, we evaluated the antenna's ability to accurately predict body-to-wall distances (BTWD) when deflected by periodic profiles. In these experiments, a servo motor rotated a rigid gear-like disk with a periodic profile (sinusoidal, triangular, or square-wave) into contact with the antenna tip, while the antenna base remained fixed (Figure \ref{fig:btwd}A). 
Ground-truth BTWD values were calculated by tracking marker positions at the antenna tip and base using video analysis. Predicted BTWD values were derived by converting measured mechanosensor capacitances into hinge angles using experimentally determined calibration functions (Section \ref{sec:quasistatic_characterization}) and subsequently performing forward kinematic calculations.

Comparing predicted and ground-truth BTWD data revealed strong agreement for all tested profiles. For the sinusoidal profile (Figure \ref{fig:btwd}B), predicted distances matched ground-truth values closely, with a maximum prediction error of approximately \SI{7.75}{\percent} occurring at extreme limits where the antenna tip momentarily slips off before reengaging into the next gear tooth. For the square-wave and triangular profiles, even lower maximum errors of \SI{4.56}{\percent} and \SI{6.34}{\percent}, respectively, were measured. These results confirm that CITRAS effectively functions as a reliable tactile sensing probe, providing accurate and repeatable distance estimation during dynamic interactions and enable robots to effectively navigate confined environments through touch alone.


\subsection{Estimating Environmental Gap Widths}
\label{sec:gap-width}

\begin{figure*}[ht!]
\centering
\includegraphics[width=1\textwidth]{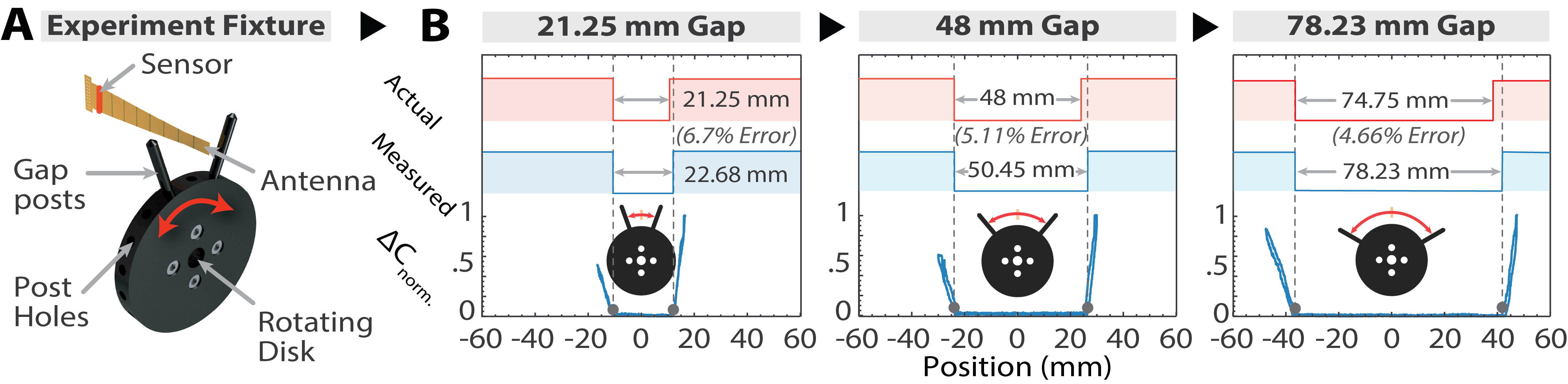}
\caption{Sizing up the environment: measuring unknown gap dimensions \textbf{(A)} Diagram of gap detection experimental setup. Three gap widths were experimentally simulated, and later estimated by post processing the fixture angle and response of sensor one. \textbf{(B)} Gap width detection results. Left to right: for increasing gap widths, raw sensor data was plotted against gap position. A sensor threshold of $3.5\ fF$ ($6\sigma$ STD above noise floor) was applied to infer the contact locations of either side of the gap, and finally compared to the ground truth gap width.}
\label{fig:gap-width}
\end{figure*}

When navigating cluttered environments, insect-scale robots frequently encounter narrow gaps that require estimation to determine navigability. Biological systems such as cockroaches employ tactile exploration strategies using their antennae to reliably assess gap dimensions prior to traversal attempts \cite{jayaram_cockroaches_2016}. 
To evaluate CITRAS's capability for similar tactile gap measurements, we simulated environmental gaps using a 3D-printed slotted disk, equipped with rigid posts of known spacing (\SI{21.25}{\milli\meter}, \SI{48}{\milli\meter}, and \SI{78.23}{\milli\meter}).
With the antenna centered between these simulated gap edges, the servo rotated to bring the gap edges sequentially into contact with the antenna tip (Figure \ref{fig:gap-width}A). Using a threshold mechanosensor capacitance change (\SI{3.5}{\femto\farad}, over six standard deviations above the noise floor) to detect gap edges, we predicted gap width from these angular measurements.
Predicted gap widths closely matched actual values: \SI{22.68}{\milli\meter} (\SI{6.73}{\percent} error), \SI{50.45}{\milli\meter} (\SI{5.11}{\percent} error), and \SI{78.23}{\milli\meter} (\SI{4.66}{\percent} error). 
 These results clearly demonstrated accurate gap width predictions for all tested cases (Figure \ref{fig:gap-width}B) and highlight CITRAS’s effectiveness in tactile gap exploration.

\subsection{Texture Discrimination}
\label{sec:texture-discrimination}

\begin{figure}[b!]
\centering
\includegraphics[width=1\linewidth]{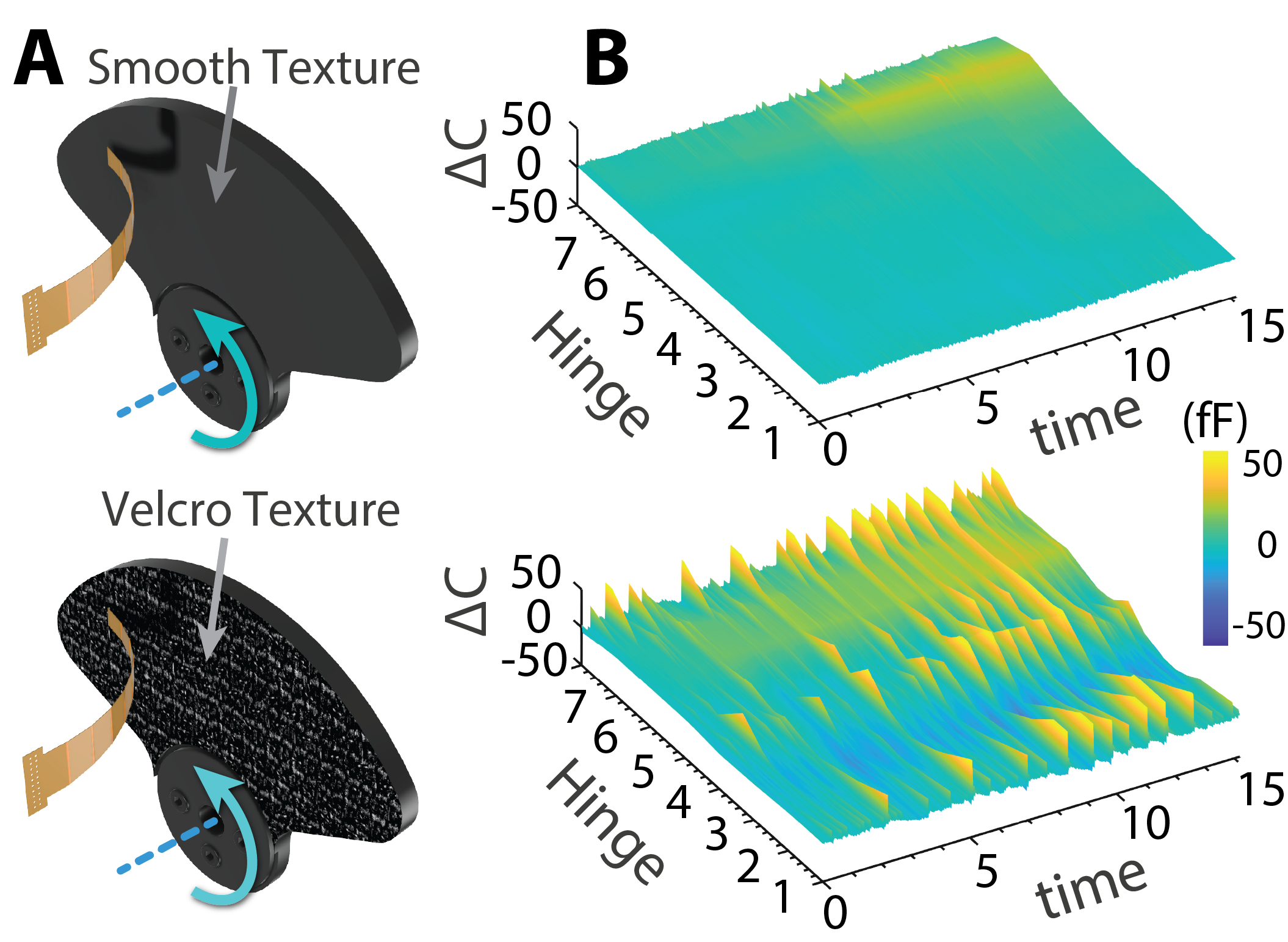}
\caption{Antenna response to variable surface roughness. \textbf{(A)} Simplified diagram of test fixtures used for texture discrimination experiments. \textbf{(B)} Resulting tactile images generated from raw sensor response. Both smooth and rough surfaces generate analogously textured tactile images.}
\label{fig:texture-discrimination}
\end{figure}

Perception of environmental textures could be critical for insect-scale robots deployed for inspection and maintenance tasks \cite{de_rivaz_inverted_2018}. To investigate CITRAS’s potential for texture discrimination, we performed tactile experiments using two distinct surface textures: a smooth surface (3D-printed PLA) and a rough surface (Velcro hooks). Each surface was mounted on a rotating servo, contacting the antenna tip at a constant linear speed of approximately \SI{2.3}{\milli\meter/\second} (Figure \ref{fig:texture-discrimination}A).
Sensor data represented as 3D tactile images (Figure \ref{fig:texture-discrimination}B) show clear differences between the two textures. The smooth PLA surface induced minimal deflections, producing small mechanosensor capacitance variations (\SI{6.2}{\femto\farad}, peak-to-peak average across hinges), whereas the rough Velcro surface caused pronounced stick-slip motion with significantly higher capacitance variations (\SI{30.1}{\femto\farad}, peak-to-peak average). Although more sophisticated measures of surface variations \cite{kabutz_integrated_2025} can be utilized depending on task specifications in future studies, these experiments demonstrate CITRAS’s potential for tactile-based material classification and texture perception.

\section{Discussion}
\label{sec:discussion}

Accurate environmental tactile perception is critical for enabling autonomous robotic navigation and exploration tasks in complex, confined spaces. While larger robotic platforms rely on sensor fusion approaches involving multiple bulky and power-intensive sensors, insect-scale robots require compact, lightweight, and power-efficient sensing solutions optimized explicitly for their SWAP constraints. CITRAS addresses these stringent requirements by directly leveraging bioinspiration from cockroach antennae, integrating mechanical compliance, distributed mechanosensing, and multifunctional tactile perception into a single laminate structure.

Our comprehensive characterization of CITRAS demonstrated robust quasistatic and dynamic sensing performance. Quasistatic tests confirmed high accuracy of mechanosensors with minimal angular prediction errors (average: \SI{0.056}{\degree}; maximum: \SI{0.795}{\degree}), enabling precise curvature determination during controlled deflections. Dynamic characterization revealed accurate tracking of rapidly changing angles (average: \SI{0.10}{\degree}), although low damping suggests future design improvements to enhance dynamic settling times through targeted material selection or structural modifications.
Beyond basic mechanical validation, CITRAS demonstrated effective multifunctional tactile capabilities, including accurate estimation of body-to-wall distances (\SI{\leq 8}{\percent} error), precise measurement of environmental gap dimensions (\SI{\leq 7}{\percent} error), and easy discrimination between smooth and rough surface textures using spatiotemporal tactile images. 

Future work will explicitly focus on advancing fabrication techniques to improve mechanical robustness. Incorporating highly viscoelastic elastomers \cite{lakes2009viscoelastic}, tunable smart materials \cite{hedrick2024femtosecond, mcclintock2021fabrication}, fluid-filled structures \cite{kellaris2021spider}, and actively controlled damping elements \cite{diller2016lightweight, hinchet2020high} could significantly improve sensor reliability and dynamic response characteristics. These enhancements would enable CITRAS-inspired antennae to more closely replicate the sophisticated tactile sensing behaviors of biological antennae, facilitating reliable autonomous navigation for insect-scale robotic platforms in real-world, cluttered environments.

\section*{Acknowledgments}

Special thanks to all the members of the Animal Inspired Movement and Robotics Lab at CU Boulder, and the Bio-Motion Systems Lab at Penn State for their support and guidance on this project. This work was supported by the Army Research Office (ARO) under grant number $W911NF-23-1-0039$ (to J.M.M. and K.J.) and National Science Foundation under CAREER grant number $2443869$ (to K.J.).

\section*{Conflict of Interest}

The authors declare no conflicts of interest in the completion of this work.

\section*{Code Availability}

All code used to generate figures and process data is available on the paper GitHub repository: \url{https://github.com/Animal-Inspired-Motion-And-Robotics-Lab/Paper-Robotic-Antenna}

\bibliography{bibliography}

\end{document}